%% file: main.tex
\documentclass[sigconf]{acmart}

\usepackage[utf8]{inputenc} %
\usepackage[T1]{fontenc}    %
\usepackage{hyperref}       %
\usepackage{url}            %
\usepackage{booktabs}       %
\usepackage{amsfonts}       %
\usepackage{nicefrac}       %
\usepackage{microtype}      %
\usepackage{natbib}
\usepackage{graphicx}
\usepackage{wrapfig}
\usepackage{amsmath}
\usepackage{bbm}
\usepackage{subcaption}
\usepackage{enumitem}

\newcommand{\PP}{\mathbb{P}}
\newcommand{\EE}{\mathbb{E}}
\newcommand{\Rd}{\mathbb{R}^d}
\newcommand{\sumn}{\sum_{i=1}^n}
\newcommand{\yif}{Y_i}
\newcommand{\propi}{e(T_i,x_i)}
\newcommand{\prop}{e(T,x)}
\newcommand{\epropi}{\hat{e}(T_i,x_i)}

\newcommand{\regikest}{\hat{m}(T, x_i)}

\AtBeginDocument{%
  \providecommand\BibTeX{{%
    \normalfont B\kern-0.5em{\scshape i\kern-0.25em b}\kern-0.8em\TeX}}}

\copyrightyear{2022} 
\acmYear{2022} 
\setcopyright{acmlicensed}\acmConference[WWW '22 Companion]{Companion Proceedings of the Web Conference 2022}{April 25--29, 2022}{Virtual Event, Lyon, France}
\acmBooktitle{Companion Proceedings of the Web Conference 2022 (WWW '22 Companion), April 25--29, 2022, Virtual Event, Lyon, France}
\acmPrice{15.00}
\acmDOI{10.1145/3487553.3524211}
\acmISBN{978-1-4503-9130-6/22/04}

\begin{document}

\title{Fair Effect Attribution in Parallel Online Experiments}

\author{Alexander Buchholz}
\email{buchhola@amazon.com}
\affiliation{%
  \institution{Amazon Music ML}
  \streetaddress{}
  \city{Berlin}
  \country{Germany}
}

\author{Vito Bellini}
\email{vitob@amazon.com}
\affiliation{%
  \institution{Amazon Music ML}
  \streetaddress{}
  \city{Berlin}
  \country{Germany}
}

\author{Giuseppe Di Benedetto}
\email{bgiusep@amazon.com}
\affiliation{%
  \institution{Amazon Music ML}
  \streetaddress{}
  \city{Berlin}
  \country{Germany}
}

\author{Yannik Stein}
\email{syannik@amazon.com}
\affiliation{%
  \institution{Amazon Music ML}
  \streetaddress{}
  \city{Berlin}
  \country{Germany}
}

\author{Matteo Ruffini}
\email{ruffinim@amazon.com}
\affiliation{%
  \institution{Amazon Music ML}
  \streetaddress{}
  \city{Berlin}
  \country{Germany}
}

\author{Fabian Moerchen}
\email{moerchen@amazon.com}
\affiliation{%
  \institution{Amazon Music ML}
  \streetaddress{}
  \city{Seattle}
  \country{USA}
}

\renewcommand{\shortauthors}{Buchholz et al.}

\begin{abstract}
  A/B tests serve the purpose of reliably identifying the effect of changes introduced in online services. 
  It is common for online platforms to run a large number of simultaneous experiments by splitting incoming user traffic randomly 
  in treatment and control groups. 
  Despite a perfect randomization between different groups, simultaneous experiments 
  can interact with each other and create a negative impact on average population outcomes such as engagement metrics. These are measured
  globally and monitored to protect overall user experience. 
  Therefore, it is crucial to measure these interaction effects and attribute their overall impact in a fair way to the respective experimenters. 
  We suggest an approach to measure and disentangle the effect of simultaneous experiments by providing 
  a cost sharing approach based on Shapley values. 
  We also provide a counterfactual perspective, that predicts shared impact based on conditional average treatment effects making use of 
  causal inference techniques. 
  We illustrate our approach in real world and synthetic data experiments. 

\end{abstract}



\maketitle

\input{intro}
\input{background}
\input{experiments}
\input{discussion}

\begin{acks}
We would like to thank Moritz von Pein, Julian Dietz, Jan Malte Lichtenberg and Matej Jakimov for their support throughout the project. 
\end{acks}

\bibliographystyle{ACM-Reference-Format}
\bibliography{bib_te}

\appendix
\newpage
\section{Appendix}
In our appendix we provide more details on causal inference as well as our experimental set-up. 

\subsection{More details on causal inference}
For the sake of completeness we introduce more involved causal inference methods that can be used in combination with our cost sharing approach. 
\paragraph{Regression adjustment}
Assuming a linear relationship between the outcome variable and pre-treatment covariates we can use a linear model to directly estimate the treatment effect. This approach, commonly used in econometrics \citep{AngristPischke2009}, has the advantage of directly providing estimates for the uncertainty of the treatment coefficients. A clear downside of this approach, however, is that the underlying assumption can be overly restrictive and a violation of the presupposed linearity can result in biased inference. 
The linear regression adjustment model is defined as 
\begin{eqnarray} \label{eq:regression_adjustment}
  \yif = \sum_{T} \alpha_T D_{i}(T) + \beta_0 + \sum_{j=1}^J \beta_j X_{i,j} + \epsilon_i, 
\end{eqnarray}
where $\epsilon_i$ is an error term, $X_{i,j}$ are our pre-treatment covariables. 
Counterfactual prediction for individual values of $K$ can be obtained using Equation \eqref{eq:regression_adjustment} for prediction of the form $\regikest \approx \EE \left[ Y_i(T) | X_i= x_i\right]$, which can be used for the computation of the conditional weighted Shapley values. An estimate for the treatment effect is obtained as $\hat{\mu}_T^{RA} = 1/n \sumn \regikest $. 

\paragraph{Self normalized IPS}
The self normalized IPS estimator makes the IPS method more robust by reducing the variance of small weights \citep{imbens2015causal, swaminathan2015self}. The corresponding estimator is biased but consistent. Its form is given as
\begin{eqnarray}
  \hat{\mu}^{snIPS}_T = \hat{\mu}_T^{IPS}/ \sumn \frac{1}{n} \frac{D_{i}(T)) }{\epropi}.
\end{eqnarray}

\paragraph{Doubly robust estimator}
The class of doubly robust estimators \citep{robins1994estimation} combines the propensity score estimation with the regression adjustment. This makes this class of estimators correct if either the propensity score adjustment or the regression adjustment is incorrect (but not both). A widely used version of this estimator is given as 
\begin{eqnarray}
  \hat{\mu}^{DR}_T = \sumn \left( \regikest + \frac{D_{i}(T)(\yif - \regikest)}{\epropi}\right)/ \sumn D_{i}(T). 
\end{eqnarray}
Here $\regikest$ are the predictions of the regression adjustment model of Equation \eqref{eq:regression_adjustment} for the treatment set to $T$ for covariate $x_i$. 

The doubly robust estimator can be estimated using a two step approach where first we estimate the multivariate propensity model by regressing the treatment on the pretreatment covariates. Then as a second step we regress the observed outcomes for treatment $T$ on the pretreatment covariates. 
The predicted outcome for the regression adjustment and the propensity score model are then used to compute the estimator $\hat{\mu}^{DR}_T$. 
This estimator has typically a higher variance than the RA estimator, if its underlying model is correct, but in practice the doubly robust property is often worth this loss.

\subsection{Details on the experiments}

\paragraph{Standard error estimation}
Confidence intervals and standard errors are computed using bootstrapping where we resample datasets 200 times with replacement (see, for example, \cite{hernan2020causal}). Then, we compute the estimated lift using the different methods introduced in Section \ref{ssec:causalinference}. 

\paragraph{Multivariate estimation of treatment effects and lift}
We estimate treatments effects and lift using a multivariate approach. Here, we estimate jointly the effect of experiment as well as their interactions. The number of treatments is $2^L-1$, where $L$ denotes the number of experiments. The empty set $\emptyset$ serves as baseline. The propensity score is derived from a multinomial model with $2^{|L|}$ different classes. 

\paragraph{Univariate estimation of treatment effects and lift}
We estimate treatments effects and the associated lift using an univariate approach looking at each experiment $l$ individually. We focus on the impact on the experiment level without taking into account potential interactions. We therefore term this approach \textit{marginal} effect estimation, as we estimate the causal effect of a binary treatment at the experiment level. 

\paragraph{Synthetic experiment}
The underlying data is generated by fixing first two parameter $\beta_1$ and $\beta_2$, then we simulate a normal distributed covariate vector $X$. Then $X$ and $\beta_1$ generate treatment assignment in a multinomial model resulting in $T$.  $X$ and $\beta_2$ are then used to add more confounding. We provide python code for the data generation below. 

\begin{verbatim}
import numpy as np
import pandas as pd
np.random.seed(42)

class GenerateSyntheticSample(object):
    """
    function that generates synthetic sample
    """
    def __init__(self, dim, m_treatments=3, rct=False, seed=None):
        self.dim = dim
        self.m_treatments = m_treatments
        self.rct = rct
        # generate random treatment effects between -1,1
        np.random.seed(42)
        self.tau = 2*np.random.uniform(
          size=(2**self.m_treatments-1))-1
        np.random.seed(None)
        self.seed = seed

        beta_1 = np.linspace(-dim, dim, num=dim)
        self.beta_1 = beta_1 / np.sum(beta_1 ** 2) ** 0.5
        beta_2 = np.linspace(dim, -dim, num=dim)
        self.beta_2 = beta_2 / np.sum(beta_2 ** 2) ** 0.5

    def generate_sample(self, n):
        if self.seed:
            np.random.seed(self.seed)

        x = np.random.normal(size=(n, self.dim))
        if self.rct:
            p = np.ones((n, self.m_treatments)) * 0.5
        else:
            p = np.zeros((n, self.m_treatments))
            for mi in range(self.m_treatments):
                p[:, mi] = 1.0 / 
                (1.0 + np.exp(-x.dot(self.beta_1)*(-1**mi)))

        d = np.random.binomial(1, p)
        treatment_strings = list(map(''.join, 
            d.astype(int).astype(str)))
        d_all = pd.get_dummies(treatment_strings, 
        drop_first=True)

        self.feature_names = d_all.columns

        y = x.dot(self.beta_2) + 
            d_all.dot(self.tau) + np.random.normal(size=n)
        weights = np.ones(n)
        return y, x, d, weights

syntheticsamplegenerator = 
  GenerateSyntheticSample(5, rct=False)
y, x, d, weights = 
  syntheticsamplegenerator.generate_sample(10000)

\end{verbatim}

\paragraph{Real world experiment}
Our real world experiment uses data from Amazon Music. 
In order to not disclose sensitive business information, we refrain from giving exact details on the experiment. 

\end{document}

%% file: intro.tex
\section{Introduction}
Randomized experiments like A/B tests \citep{siroker2013b}, also known as randomized control trials (RCT), identify the effect of a treatment on a population. They are omnipresent on online platforms \cite{kohavi2009controlled, tang2010overlapping}. 
Analyzing their outcomes allows to 
measure individual and interaction effects \citep{montgomery2017design} for a large set of treatments in situations where treatments are perfectly randomized or potentially confounded \citep{blackwell2020noncompliance}.
However, massive simultaneous experimentation can have a cost for overall engagement metrics that the platform provider monitors. We provide an approach for measuring and disentangling the impact of parallel experiments and attributing their respective cost. 

If several experiments are run on an online platform (for example, promoting new music content or showing several ads on a search engine), the overall user engagement measured across all experiments might decrease as users get overwhelmed. 
As an illustrative engagement metric we use throughout the click-through-rate (CTR). 
In this situation it is straightforward to measure the impact of each individual experiment on their treatment and control group, but how the experiments contribute to the average engagement across all users is unclear. 
In what follows we will assign costs $\Delta_l$ to individual experiments $l \in L$ relative to their impact by decomposing the overall loss in engagement due to the set of all active experiments $L$. Using concepts from cost sharing in game theory we can think about an experiment as a player. Several active players form a coalition, i.e., the specific treatment combination that a user gets exposed to. 

We are interested in attributing the effect of an experiment to the overall population outcome, meaning that each player gets an assigned cost for being part of a coalition. %
We want to assess the loss compared with the baseline $\mu_0$ that corresponds to the absence of experimentation (i.e., the control group). 
The overall loss is then denoted by $\Delta = \bar{Y}-\mu_0$ and we provide a decomposition such that $\Delta = \sum_{l \in L} \Delta_l$ where different experiments might interact with each other. This question is crucial if we want to check overall effects by experimentation, limit impact on the user experience and attribute the footprint fairly to stakeholders that run experiments. 
Our notion of fairness is derived from the concept of Shapley values, see \cite{moulin2004fair}. 

\paragraph{Contributions}
Our contributions are the following. 
\begin{itemize}
  
  \item We suggest two approaches based on causal inference and cost sharing games to attribute the impact of interacting treatments to average population level outcomes. We illustrate that a marginal perspective, looking at individual experiments only, would lead to misleading conclusions. 
  \item We demonstrate the effectiveness of our approach in a large online user study on the Amazon Music platform as well as in a controlled, synthetic setting. 
  \item We suggest an extension to conditional cost sharing that allows to predict shared impact in a counterfactual setting. 
\end{itemize}

The rest of this paper is structured as follows. In Section \ref{sec:related_work} we review related work. Section \ref{sec:background} introduces background material and our suggested methodology. Section \ref{sec:experiments} details the experimental setup and highlights our results. Finally, Section \ref{sec:conc} discusses our results and concludes. 

\begin{table}[h]
  \begin{tabular}{cccc}
    \toprule
    Notation&Definition&Notation&Definition\\
    \midrule
    $Y_i$ & Outcome & $\bar{Y}$ & Population average \\
    $D_i(T)$& Treatment indicator & $T_i$& Treatment\\
    $\mu_T$ & Average effect & $\Delta$& Total cost\\
    $\Delta_l$ & Attributed cost& $\propi$ & Propensity\\
    $X_i$ & Covariates& $l$ & Active experiment\\
  \bottomrule
\end{tabular}
\caption{Notation}
\label{tab:notation}
\vspace*{-10mm}
\end{table}

\section{Related Work} \label{sec:related_work}
Our work can be divided along three lines: (a) causal inference approaches for multivalued treatments; (b) measurement of the efficiency of ads on platform business, approaches to study parallel experiments and cost sharing games in game theory; and finally (c) trade-offs in recommender systems and multi-sided platforms.

\paragraph{Multiple simultaneous treatments in causal inference}
Studying the effect of multivalued treatments goes back to at least the seminal work of \cite{Imai2004} on the generalized propensity score. 
Estimating multivalued treatments effects \citep{Feng2012, Linden2016, Nian2019, Hu2020} has since then received substantial attention in the field of econometrics and biostatistics \citep{Mccaffrey2013,Lopez2017}. Another perspective on analyzing simultaneous treatments is using experimental design approaches, see \cite{cox2000theory, montgomery2017design}. %
In our work we make extensive use of this methodology to quantify the impact of experiments. 

\paragraph{Measuring of ad effectiveness and cost sharing games}
Due to its economic importance, the measurement of ad effectiveness has become a major field of application for causal inference, see for example recent work by \cite{Gordon2019, Lada2019} and \cite{Lin2019parallel} for a perspective on parallel experimentation. We take a holistic perspective on the problem as we want to measure and disentangle combined impact of several experiments such as showing ads and promotions. We assess the impact of the experiments on the combined user experience which could be contrary to the aims of the experiment providers. 
Cost sharing games assign a value to contributing players and go back to the introduction of Shapley values by \cite{shapley1953value}. See also \cite{balkanski2015mechanisms,balkanski2017statistical, jain_mahdian_2007} for more background. Attributing treatment effects to various marketing channels in online advertising has been approached by combining cost sharing approaches with causal inference, see \cite{singal2019shapley}. Our approach is distinct from the marketing attribution problem due to the parallel nature of the experiments. %

\paragraph{Trade-offs in recommender systems} 
Optimizing engagement along other business objectives 
has become a crucial topic for multi stakeholder recommender systems, 
\citep{Abdollahpouri17, Stanton2019}. Typical applications consist in automatic allocation of sponsored search \citep{Malthouse2019, Zhao2020} or multi-relevance ranking \citep{Momma2019} employing techniques from constraint optimization. Multi-objective optimization in market places \citep{mehrotra2020bandit, Ding2019, Lin2019} has also seen growing interest. For a recent line of work on the multi-sided and multi-objective nature of online platforms see, e.g., \cite{Burke2016,Burke2017,Nguyen2017,Mehrotra2018}. In what follows we provide a perspective that splits impact fairly between different stakeholders and thus contributes to the understanding of platforms and their multi-sided nature. 

%% file: background.tex
\section{Background and Suggested Approach} \label{sec:background}

Our aim is to decompose the average observed outcome $\bar{Y}$ among the 
different parties $l$ that run experiments. We want to achieve this both in 
an empirical and counterfactual fashion. We consider the case of RCTs, where exposure to treatments is perfectly randomized, and observational studies, where treatments are potentially confounded. We employ the potential outcomes framework by \cite{rubin1974estimating}. For a recent survey see \cite{yao2020survey} or as major reference \cite{imbens2015causal}.

\subsection{Causal Inference Techniques} \label{ssec:causalinference}
We introduce required notation and the methods that we use to identify treatment effects.

\paragraph{Notation} We denote our outcome variable as $Y_i(T_{i})$, where every unit of observation $i$ can receive $2^{|L|}$ different treatments $T_{i} \in \mathcal{P}(L)$, which informs which treatment observation $i$ receives, and $\mathcal{P}(L)$ is the power set denoting active experiments $l \in L$. 
Hence the experiments $l \in T$ inform us which experiment is part of a treatment. 
The empty set $\emptyset$ corresponds to the baseline $\mu_0$, namely the control group. 
Pre-treatment covariables (i.e., measured before treatment is assigned) are denoted $X_i \in \Rd$.
We define a treatment indicator $D_i(T)$ as
$$
  D_i(T) =
  \begin{cases}
    1, ~\text{if unit }i\text{ receives treatment }T \\
    0, ~\text{otherwise}
  \end{cases}.
  $$
The observed outcome $\yif$ is then written 
as
$
  \yif = \sum_{T \in \mathcal{P}(L)} Y_{i,T}D_i(T),
$
using the shorthand $Y_{i,T} = Y_i(T_i)$. 
We denote the expected outcome as 
$
  \mu_T = \EE \left[ Y_{i,T} \right], 
$
and we are interested in 
the population average treatment effect given as 
$
  \text{ATE}_{T,0} = \mu_{T} -\mu_{0}, 
$
or more generally treatment comparisons of the form 
$\text{ATE}_{T,S} = \mu_{T} - \mu_{S}$. 
We define the lift over the baseline (in \%) as 
$
  \text{lift}_{T,0} = \text{ATE}_{T,0}/\mu_0 \times 100. 
$

\paragraph{Assumptions}
Following the definition in \cite{yao2020survey} the usual identification assumptions are 
\begin{itemize}
  \item Stable unit treatment value assumption (SUTVA): the potential outcome for a unit does not vary with treatments assigned to other units. There are no different versions of the treatment. 
  \item Ignorability: Given the covariables $X_i$, the treatment assignment $T_i$ is independent of the potential outcome, i.e., $T_i \perp Y_i(T_i) | X_i $
  \item Positivity: for any value of $X$, treatment is stochastic, i.e. $\PP(T_i | X_i = x_i) > 0~\forall~T_i, x_i$. 
\end{itemize}

\paragraph{Propensity modeling}
At the heart of most techniques in causal inference lies the propensity score \citep{rosenbaum1983central} that is defined as the probability of receiving treatment: 
\begin{eqnarray} \label{eq:propensity}
  \propi = \PP(T_i| X_i = x_i).
\end{eqnarray}
The propensity scores quantifies the fact that receiving treatment might depend on characteristics of the observation units. 
The propensity score can be modeled using, e.g., a multinomial logistic regression or non-parametric models. We denote the estimated propensity score by $\epropi$. 

\paragraph{Mean treatment}
As a first approach for measuring treatment effects we introduce an estimator based on population averages. 
In the case of RCTs %
the average outcome can  be estimated as 
\begin{eqnarray} 
  \hat{\mu}_T^{mean} = \sumn D_i(T) \yif / \sumn D_i(T). 
\end{eqnarray} 

\paragraph{Inverse propensity weighting}
As a second approach for estimating treatment effects under confounding, we use inverse propensity score weighting \citep{hirano2003efficient}. Under the ignorability of treatment assumption we estimate the expected outcome of treatment $T$ by
\begin{eqnarray} \label{eq:ipsestimator}
  \hat{\mu}_T^{IPS} = \frac{1}{n}\sumn \frac{D_i(T) \yif}{\epropi}. 
\end{eqnarray}

\paragraph{Estimation of treatment effects}
We estimate the average treatment effect over the baseline as 
$
  \hat{ATE}^{\bullet}_{T,0} = \hat{\mu}_T^{\bullet} - \hat{\mu}_0^{\bullet},
$
where $\bullet$ denotes the methods \{mean, IPS\} as defined above. 
For comparing the effect of treatment $T$ over $S$, we use $\hat{ATE}^{\bullet}_{T,S}$. 

\paragraph{Marginal effects}
Our exposition so far models the potential interaction of all experiments. We also estimate marginal effects, that ignore other running experiments. They are obtained using binary treatments of the form $T_{i,l} \in \{0,1\}$.

\subsection{Cost Sharing Games And Our Suggested Approach} \label{ssec:attribution}

The introduced approaches estimate the impact of different experiment combinations, but do not assign the individual contribution of $l$ to the treatment $T$. The goal of our paper is to share the total impact over the baseline, i.e. 
$ \mathbb{E}[\bar{Y}] - \mu_0$, among all contributors. %
This problem is known in the field of game theory as cost sharing game. Transcribed to our setting, the term \textit{player} corresponds to active experiments. Players form \textit{coalitions} (i.e., specific treatments). The term \textit{grand coalition} denotes the set of all experiments. 
In a cooperative game with $|L|$ players each player $l \in L$ is assigned a value $\phi_l(v)$ for the game $v$. If players form a coalition $S \subset L$ this results in a cost of the coalition $v(S)$. 
Cost sharing mechanism should satisfy the following (axiomatic) properties for the assigned cost $\phi_l(v)$: 
\begin{enumerate}[label=(\alph*)]
  \item Symmetry: it does not matter in which order the players $l$ are numbered.
  \item Balanced budget: $\sum_{l \in L} \phi_l(v) = v(L)$ the sum of the individual values should equal the total outcome. 
  \item Null player: $\phi_l(v) = 0$ if $l = \emptyset$, a player that does not contribute to the value of the game must have a null contribution.
  \item Additivity: for two games $v,w$ $\phi_l(v) + \phi_l(w) = \phi_l(v+w)$. 
\end{enumerate}
The only cost sharing mechanism that satisfies all of the above criteria is the Shapley value \citep{shapley1953value}. See \cite{moulin2004fair} for its connection with fairness and distributive justice. It is defined as
$$
\phi^L_l(v) = \sum_{S \subset L - l} \frac{|S|! (|L|-|S|-1)!}{|L|!} [v(S \cup l)-v(S)], 
$$
where we make explicit the dependence on the grand coalition $L$.  
One distinctive feature of Shapley values is the use of marginal contributions $v(S \cup l)-v(S)$. 
We now suggest two approaches to construct decompositions of the form 
$
  \mathbb{E}[\bar{Y}] - \mu_0 = \sum_{l \in L} \Delta_l, 
$
that satisfy the balanced budget condition (b) as well as a perspective conditional on covariate values. 
The expectation of the average observed outcome is decomposed in a weighted sum of contributions:
$\mathbb{E}[\bar{Y}] = \sum_{T \in \mathcal{P}(L)} \mu_T \times \mathbb{P}(T),$
where $\mu_T$ and $\mathbb{P}(T) := \EE_X [e(T,X)] $ can be estimated using the techniques introduced before. The cost of a coalition is given as 
$v(S) = \mu_{S}-\mu_0$. 

\paragraph{Weighted Shapley cost sharing}
We suggest weighted Shapley cost sharing as solution to the cost sharing problem. Our solution is 
based on the concept of Shapley values that takes into account that not all experiments are active in parallel. The Shapley value conditional on the treatment $T$ is $\phi^T_l(v)$, meaning that we only consider the subsets of coalitions up to $T$ for computing this value. A weighted decomposition is then given as 
\begin{eqnarray}
  \tilde{\Delta}_l = \sum_{T \in \mathcal{P}(L)} \mathbb{P}(T) \phi^T_l(v).
\end{eqnarray}
It is easily shown that weighted Shapley values are budget balanced.
Note that if $l \notin T$, the null player property guarantees that $\phi^T_l(v)=0$. Thus, this approach inherits all the favorable properties of Shapley values. 

\paragraph{Weighted average cost sharing}
As an alternative approach we suggest to divide the impact of treatment $T$ equally among individual experiments $l$ contributing to $T$, an approach corresponding to average cost sharing \citep{tauman1988aumann}.
If a specific experiment $l$ is part of a treatment, i.e. $l \in T$, we compute the impact of $l$ on the total outcome via
\begin{eqnarray}
  \Delta_l = \sum_{T \in \mathcal{P}(L)} [\mu_{T} - \mu_0] \times \mathbb{P}(T) \frac{\mathbbm{1}\{l \in T \}}{ |T|}, %
\end{eqnarray}
where $|T|$ is the number of active experiments inside the treatment $T$. The impact of $l$ is thus the weighted impact over the baseline. 
The suggested decomposition has the balanced budget property but has the inconvenience that null players are not necessarily ignored. This makes this approach potentially unfair compared with weighted Shapley cost sharing. 
See also \citep{jain_mahdian_2007, balkanski2017statistical} for further cost sharing and estimation approaches. 

\paragraph{Conditional weighted cost sharing}
As an extension to the two approaches we suggest a conditional perspective, where we look at conditional average treatment effects (CATE) \citep{abrevaya2015estimating} of the form
$\mu_T(x) = \EE \left[ Y_i(T_i) | X = x \right]$. 
This idea provides insights on the impact for subgroups of the population. 
Estimating the CATE requires a model that predicts the outcome at the given covariate value $X=x$. In combination with the propensity score $\prop$ we then define conditional cost sharing of the form $\Delta(x) = \sum_{l \in L} \Delta_l(x)$,  
which quantifies the impact on a specific subgroup of the population.
The conditional weighted cost sharing approach can both be used in combination with Shapley or average cost sharing. 

\paragraph{Approximating Shapley values}
Our discussion so far assumed that the value of all possible coalitions are observed. In the case of missing combinations approximation techniques such as \cite{maleki2013bounding} could be used. The same idea can be used if the number of parallel experiments gets large, since 
the exact computation of the Shapley value has a exponential runtime which can be prohibitive. We leave the investigation of this for future work. 

%% file: experiments.tex
\section{Experiments and Results} \label{sec:experiments}

\begin{figure}[h]
    \centering
    \includegraphics[width=0.8\linewidth]{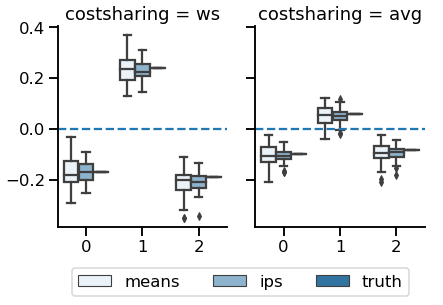}
    \caption{Shapley cost sharing (left) and average cost sharing (right) in a synthetic experiment. The shared cost is shown on the y-axis, the attributed experiment (numbered by 0,1,2) is indicated by the x-axis. The data has been simulated under covariate confounding which explains the higher error of the means estimator. The average cost sharing disentangles the contributions of the experiments less. The ground truth is indicated by a straight line next to the boxplots. Average cost sharing waters the contributions down to 0 and hence differences in the impact are not properly attributed.}
    \label{fig:results_se}  
    \vspace*{-5mm}
  \end{figure}

\paragraph{Synthetic experiment}
We illustrate the difference between weighted Shapley cost sharing and average cost sharing in a synthetic experiment. See the Appendix for more details. 
In this experiment we deliberately introduce confounding by making treatment assignment dependent on the covariates $X_i$. Figure \ref{fig:results_se} illustrates the result: the approach based on IPS weighting, has less variance and yields more precise estimates compared to a naive strategy based on sample means. The approach based on average cost sharing results in all attributed costs watered down towards $0$ and makes detecting significance more difficult. The weighted Shapley cost, however, allows a clear disentanglement when combined with IPS weighting.

\begin{figure}[h]
    \centering
    \includegraphics[width=0.75\linewidth]{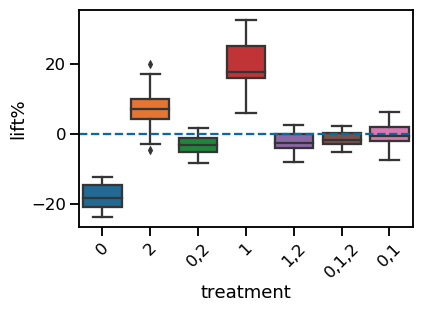}
    \caption{Impact of all experiment combinations (treatment, shown on the x-axis). We show the lift relative to the baseline in \% (y-axis). The first experiment alone had a substantial negative impact, whereas the second experiment alone led to an uplift in engagement.}
    \label{fig:multivariate_treatments}
    \vspace*{-4mm}
  \end{figure}

\paragraph{Real world experiment}
As a real world experiment we show results obtained on the Amazon Music platform
using millions of observations. 
We studied three different experiments that were run in parallel for Amazon Music users in 7/2021 over two weeks. The experiments consisted in showing new editorial 
music content to a subset of users where some users saw the respective treatment and others the control. Treatments were perfectly randomized (checked using covariate balance). 
Running several experiments at once led to a reduction in $-1.27\%$ of overall CTR compared with no exposure to the three experiments. However, the contributions to the overall reduction was 
mostly due to a single experiment as Figure \ref{fig:multivariate_treatments} illustrates. 
Experiment 0 lead to a loss on a small group, though this effect disappears 
when the experiment is active alongside the two other experiments (1,2). 
This interaction is not identified, when looking at marginal contributions only (see Table \ref{tab:results_rw}), where a negative impact is indicated, 
but not significant for any of the experiments. This is due to a dominance of the treatment where all experiments are active (around $66\%$ of the population). 
Using an average cost sharing approach does not solve the problem, as all contributions are watered down to towards $0$, without a significant indication of negative impact. 
The only approach that draws a clear picture is the one based on weighted Shapley values. Here, experiment 0 is significantly negative, whereas experiment 1 has a significantly positive lift on the overall experience (see Table \ref{tab:results_rw}). 
As experiment 0 is negatively contributing to the overall user experience, a resulting decision would be to disable it.

\begin{table}
    \caption{Shared cost and marginal impact for the experiments on Amazon Music. Values measured as lift in \%, the parentheses contain $95\%$ confidence intervals. Bold values indicate significance. }
    \label{tab:results_rw}  
\begin{tabular}{llll}
    \toprule
     &                        Exp. 0 &                     Exp. 1 &          Exp. 2 \\
    \midrule
    Average cost      &     -0.59  &  -0.24 &   -0.44\\
    (\%)      &     (-1.35, 0.21) &  (-0.97, 0.64) &   (-1.26, 0.32) \\
    Marginal Impact&     -1.33 &   -0.87 &  -1.28 \\
    (\%) &      (-3.56, 1.20) & (-3.12, 1.85) & (-3.56, 1.31) \\
    Shapley cost       &  \textbf{-10.19} &    \textbf{7.93}&   0.99\\
    (\%)   &   \textbf{(-12.31, -7.90)} &    \textbf{(4.45, 10.97)} &    (-1.41, 2.78) \\
    \bottomrule
    \end{tabular}
    \vspace*{-3mm}
\end{table}

%% file: discussion.tex
\section{Discussion and Conclusion} \label{sec:conc}

We showed how to use causal inference techniques and cost sharing approaches to estimate and disentangle the effect of parallel experiments. 
Our weighted Shapley value approach attributes impact of parallel experiments and is a step necessary towards adaptive experimentation 
that limits impact beyond a given budget per experiment. The causal nature allows to predict impact on subgroups and hence 
creates a more fine grained perspective that goes beyond average impact. As future avenue we see the investigation of approximate Shapley values and missing data issues that arise if not all experiment combinations have materialized. We see benefit in investigating the combination with latest causal inference techniques such as doubly robust or double ML methods \cite{bang2005doubly, chernozhukov2017double}.